\documentclass[conference]{IEEEtran}
\IEEEoverridecommandlockouts
\usepackage{cite}
\usepackage{amsmath,amssymb,amsfonts}
\usepackage{algorithmic}
\usepackage{graphicx}
\usepackage{textcomp}
\usepackage{xcolor}
\def\BibTeX{{\rm B\kern-.05em{\sc i\kern-.025em b}\kern-.08em
    T\kern-.1667em\lower.7ex\hbox{E}\kern-.125emX}}
\begin{document}

\title{English Out-of-Vocabulary Lexical Evaluation Task}

\author{\IEEEauthorblockN{1\textsuperscript{st} Han Wang}
\IEEEauthorblockA{\textit{Dept. of Computer Science and Engineering} \\
\textit{Texas A\&M University}\\
College Station, Texas, USA\\
hanwang@tamu.edu}
\and
\IEEEauthorblockN{2\textsuperscript{nd} Ye Wang}
\IEEEauthorblockA{\textit{Dept. of Electrical and Computer Engineering}\\
\textit{Texas A\&M University}\\
College Station, Texas, USA\\
wangye0523@tamu.edu}
\and
\IEEEauthorblockN{3\textsuperscript{rd} Xinxiang Zhang}
\IEEEauthorblockA{\textit{Dept. of Electrical Engineering}\\
\textit{Southern Methodist University}\\
Dallas, Texas, USA\\
xinxiang@smu.edu}
\and
\IEEEauthorblockN{4\textsuperscript{th} Mi Lu}
\IEEEauthorblockA{\textit{Dept. of Electrical and Computer Engineering} \\
\textit{Texas A\&M University}\\
College Station, Texas, USA \\
mlu@ece.tamu.edu}
\and
\IEEEauthorblockN{5\textsuperscript{th} Yoonsuck Choe}
\IEEEauthorblockA{\textit{Dept. of Computer Science and Engineering} \\
\textit{Texas A\&M University}\\
College Station, Texas, USA\\
choe@tamu.edu}
\and
\IEEEauthorblockN{6\textsuperscript{th} Jingjing Cao}
\IEEEauthorblockA{\textit{School of Logistics Engineering}\\
\textit{Wuhan University of Technology}\\
Wuhan, China\\
bettycao@whut.edu.cn}
}

\maketitle

\begin{abstract}
Unlike previous unknown nouns tagging task, this is the first attempt to focus on out-of-vocabulary (OOV) lexical evaluation tasks that does not require any prior knowledge. The OOV words are words that only appear in test samples. The goal of tasks is to provide solutions for OOV lexical classification and predication. The tasks require annotators to conclude the attributes of the OOV words based on their related contexts. Then, we utilize unsupervised word embedding methods such as Word2Vec and Word2GM to perform the baseline experiments on the categorical classification task and OOV words attribute prediction tasks.
\end{abstract}

\begin{IEEEkeywords}
word embedding, Gaussian mixture, lexical tagging
\end{IEEEkeywords}
\vspace{12pt}

\section{Introduction}
\vspace{12pt}

The evolution of modern English language brings new words in and eliminates old words out. Thus out-of-vocabulary (OOV) handling is an inevitable challenge among nearly all natural language processing topics: In cross-lingual translation, the quality of translation is heavily dependent on the identification of OOV words. In speech-to-text translation, detecting and modelling the OOV words can significantly improve the lexical completeness of the input. In semantic analysis, a majority of OOV words are proper names (PN) which are important for discovering the contextual concept relatedness. In social network sentiment analysis, new words are created every day on the Internet, thus the ability of handling OOV words can also contribute to the robustness of the model. Moreover, the concept of OOV handling is closely related with the emerging "zero shot learning (ZSL)" \cite{zhang2015zero} in other machine learning studies.

Therefore, it is indispensable to investigate the metrics of evaluating the OOV classifications and predictions. Although in Natural Language Processing (NLP) study there are various semantic evaluation tasks for computational semantic analysis such as \cite{wang2016self,wang2017combining,siddharthan2006syntactic,zhang2016effective,Yin2019SpatialTemporal}, the majority of current evaluation tasks focus on word sense disambiguation and text simplification. The authors in \cite{dagan2006pascal} and \cite{wang2017combining} describe their work to capture major semantic inferences across applications. Finding a valid substitution with the given context has been proven to be effective in question answering, abstract extracting, etc. The task proposed in this paper concentrates more on the dominant OOV semantic prediction. The main difference between the other semantic simplification tasks is that OOV words may not impede the performance in their tasks, while in our proposed task, the outcome is directly related with the OOV words.   

In previous unknown nouns supersense tagging, \cite{curran2005supersense} and  \cite{ciaramita2003supersense}, rule-based models were proposed within the scope of WordNet. The main limitations are:
\begin{itemize}
    \vspace{12pt}

    \item The models are heavily dependent on some particular features of the unknown words such as the part of speech tagging and chunking, morphological analysis and even gramatical relation extraction.
    \vspace{12pt}

    \item Supersense method is based on WordNet hierarchy, called \textit{lexicographer files}, which contains redundant prior knowledge.
    \vspace{12pt}

    \item These models cannot handle words that do not exist in WordNet. However, the vocabulary of Wikipedia we used is ten times larger than that of WordNet.
    \vspace{12pt}

\end{itemize}

The task proposed in this paper jumps out from the above limitations. First, both training and testing data we utilized are from Wikipedia, which require no prior knowledge. Second, the baseline experiments are purely unsupervised and are based on vector-space rather than predefined rules. 

Word Sense Disambiguation (WSD) is to identify the meanings of ambiguous words. We apply WSD in our baseline experiments and compare it with non-WSD models.   
\vspace{12pt}

\section{Related Work}
\vspace{12pt}

The author in \cite{ciaramita2003supersense} proposed supersense to classify common nouns that extends named entity classification. Based on supersense, another author in \cite{curran2005supersense} improved the rule-based model by adding hand-coded unseen nouns tagging within the scope of WordNet. 
\cite{biran2011putting} built the text simplification system closely related with WordNet to obtain the word pairs as hypernym or synonym. The author in \cite{specia2012semeval} compiled the corpus based on the lexical substitution at SemEval-2007 with manually annotation. However, OOV prediction task is similar with lexical simplification whilst the difference is obvious. First, it is required to acquire prior knowledge as much as better for text simplification. Few OOV words in our task exist in WordNet, not to mention the creation of the thesaurus. With a never-seen word, the key to understand it is to familiarize with the given context. Second, most of the test words and the candidates in lexical substitution tasks such as \cite{mccarthy2007semeval} are daily words. Thus, the demand for comprehension related to the whole context is less essential than ours.     

OOV prediction obstructs the text representation because learning representation in word is the cornerstone for the further text understanding in human sense. In the past, one-hot encoding \cite{manning2008boolean} has been used for the word indexing because of the simplicity. Nevertheless, the limitation of the one-hot encoding is apparent. This algorithm produced a sparse matrix for each word representation, but the computational power is much weak, which means it is impossible for the applications with big data. Besides, this method cannot extract the relationship between words, let alone the sentences. Compared with one-hot encoding, word embedding \cite{rumelhart1986david} and \cite{bengio2003neural} utilizes semantic and syntactic information, which can extract more relationship than one-hot encoding. 

However, word embedding also has been constrained by the poor hardware development in previous decades and the algorithm's high time complexity. Recently, \cite{mikolov2013distributed} proposed two models (skip-gram and continuous bag-of-words) named Word2Vec for effective learning representation. In addition, those methods succeed to represent word from sparse vectors into dense vectors, which favored for next training or clustering. \cite{wang2017comparisons} compared the detailed performance between sparse and dense vectors in short text classification.
  
Those models mentioned above suffers a common weakness, each word is mapped to a unique vector. In fact, numerous words can represent multiple senses in different conversations. Consequently, \cite{reisinger2010multi} proposed multi-prototype model for the polysemy. After that, other related models \cite{huang2012improving,tian2014probabilistic,vilnis2014word,wang2019attention,yin2019DLRG,,wang2018optimized,wu2019accurate,Zhang2019Visual} have been inspired such as probabilistic model, Gaussian and Gaussian mixture models and so on. 

\vspace{12pt}

\section{Task Statement}
\vspace{12pt}

OOV handling in natural language processing is still a sophisticated task due to several reasons. Lack of definition or explanation about the particular OOV word is one of the obstacles to accurately predict the meanings of OOV words. Another aspect is, newly created words usually have a comparably lower frequency than normal words, and sometimes they are eliminated when the minimal count is performed in preprocessing, Therefore, recognition of OOV highly depends on the quality of corpora. Given the fact that no corpus in use is the latest real-world corpus, it is essentially helpful to take the OOV classification and prediction capacity into consideration as a robust metric of NLP models. We hence propose our OOV classification and prediction tasks. Each task contains three parts, an OOV word, context and attributes. 
The given context is as close as a description to the OOV word and it can be either long or short regarding the rarity and difficulty of the OOV word. 
We assigned several attributes to each of the OOV word based on the corresponding context. The attributes can be the topic of the context, the potential or prior characteristics about the OOV word. Attributes are provided by independent annotators. We invited five individuals in USA, they are pursuing master or PhD degrees and are all fluent but non-native English speakers. Each context is concluded by three random annotators and we intersect the common attributes. 

Another contribution in this paper is that we introduced a category classification task of OOV words with context. When we analyzed the rare words in Wikipedia English corpus, we found that many of the words are chemical and medical science terminologies with obscure contexts, which could not be comprehensively understood by human annotators. In order to make the proposed task more generic, we discarded those contexts, and hand-picked 5 categories: Greek mythology, locations, animals, plants and technology. 

\subsection{Data source}

The corpus was collected from Wikipedia dumps\\ (\url{https://dumps.wikimedia.org/backup-index.html}). The raw XML format corpus contains various types of pre-defined categories, each category has an unambiguous structure tree with its children nodes representing the sub-categories, which is similar to WordNet (\url{https://wordnet.princeton.edu/}). Wikipedia category tree has much more nodes compared with WordNet. The majority of the selected OOV words actually do not exist in WordNet. 

\subsection{Data selection}
Regarding the rules of OOV sampling, firstly words with both too high and too low occurrence were eliminated because we need to limit the data size and those eliminated words are usually less informative, which can be treated as stopwords. A reasonable data size is vital in terms of performance and speed optimization for both training and testing. Secondly, an OOV word is restricted to appear only in given context, and this is also our condition to retrieve such an OOV. An advantage to utilize Wikipedia as the data source is its sufficient entries satisfying the requirements.

After the occurrence filtering, there yet exists numerous OOV candidates. Thus we only select some particular but convictive categories. As mentioned above, according to statistical analysis, those five categories contain abundant OOV words and their given contexts are readable. Oppositely, in the category of medicine and chemistry, the context is not readily comprehensible for our annotators.    

The last step for data selection is the attribute extraction. Each OOV word and its context is designated to three random annotators. After the individual conclusion, an attribute intersection is performed to acquire final results. 

\textbf{Examples}:

Word: \textit{arachis}

Context: 

\textit{Arachis is a genus of about 70 species of annual and perennial \textbf{flowering} plants in the \textbf{pea} family (\textbf{Fabaceae}), native to South America, and was recently assigned to the informal monophyletic Pterocarpus clade of the Dalbergieae.}  

This OOV's category is plant, while the attributes are pea, flower and fabaceae, which can all be found in context.

Word: \textit{winwebsec}

Context: 

\textit{Winwebsec is a category of \textbf{malware} that targets the users of Windows operating systems and produces fake claims as genuine anti-malware software, then demands payment to provide fixes to fictitious problems.}

This OOV's category is technology, while its attributes are malware, adware, spyware, and the context contains only malware.   

The whole data set for the two tasks can be accessed from (\url{https://github.com/hwangtamu/OOVLexical})

\subsection{Task 1}
The first task is to test if the positive examples are correctly classified and the negative examples are correctly excluded from their corresponding categories. More specifically, the ground truth is set as the higher level Wikipedia category names from which the entries were taken. 
\begin{equation}
S_1 = \frac{1}{N}\sum_{i=1}^{N}{\min{R_i}}
\end{equation}
where $R_i$ represents rank of a correct prediction and $N$ is the total number of test samples.

\subsection{Task 2}
The second task is to test if top $K$ semantic predictions of the OOV words hit the human annotated attributes. Each OOV word can have up to 5 annotated share-weighted attributes, therefore we proposed a different scoring criterion than Task 1. 
\begin{equation}
S_2 = \frac{1}{KN}\sum_{i=1}^{N}{bool( {\exists{w_{ij}}\in W_i \cap \hat{W_i}}})
\end{equation}
where $W_i$ is the top $K$ prediction set of the $i$th test sample and $\hat{W_i}$ is the annotated set of $i$th test sample.

We provide details of baseline experiments in next chapters, with the $S_1$ and $S_2$ scores reported. 
\vspace{12pt}

\section{Experiment}
\vspace{12pt}

We used Word2Vec \cite{mikolov2013distributed} and Word2GM \cite{athiwilson2017} as baseline models of the tasks. The data set we utilized for training is a subset of the Wikipedia corpus with $\sim{37}$M tokens, and $\sim{340}$K unique tokens. The low frequency words (appear less than 5 times) are removed from the training set, leaving a vocabulary of $\sim{75}$K to out models. 

\subsection{Word2Vec}
The skip-gram Word2Vec model can efficiently project a vocabulary of words into a finite dimensional vector space $R^d$ by maximizing
\begin{equation}
P(c|w;v_{c_i}, v_{w_i}) = \frac{e^{v_c\cdot v_w}}{\sum_{c'\in C}e^{v_c'\cdot v_w}}
\end{equation}
where $v_c$ and $v_w\in R^d$ are vector representations for context $c$ and word $w$ respectively \cite{goldberg2014word2vec}.

One limitation of Word2Vec is that each word can only be represented as one single dot in $R^d$, and many English words have multiple meanings in use.

The Word2Vec model we trained has 50 dimensions, and in order to compare the performance, we also used a pretrained Word2Vec model from Facebook's FastText (\url{https://fasttext.cc/docs/en/english-vectors.html}). It was trained with 16B tokens, has 300 dimensions and contains 1M most frequent words.

\subsection{Word2GM}
As an attempt to solve the limitation of Word2Vec mentioned above, the word embedding with Gaussian mixtures (Word2GM) \cite{athiwilson2017} tried to represent each word with a Gaussian mixture in high-dimensional space $R^d$:
\begin{equation}
f_w(\overrightarrow{x})=\sum_{i=1}^{K}{p_{w,i}\mathcal{N}(\overrightarrow{x};\overrightarrow{\mu}_{w,i};\overrightarrow{\Sigma}_{w,i})}
\end{equation}
The objective function, derived from Word2Gauss \cite{vilnis2014word} is to minimize the max-margin ranking:
\begin{equation}
\begin{split}
L_\theta(w, c, c') = &\max(0, m - \log E_\theta(w, c) \\ 
&+ \log E_\theta(w, c'))
\end{split}
\end{equation}
where the term $c'$ is from negative sampling. Several metrics can be applied to measure the similarity of words:

\begin{figure}
  \centering
    
      \includegraphics[width=0.5\textwidth]{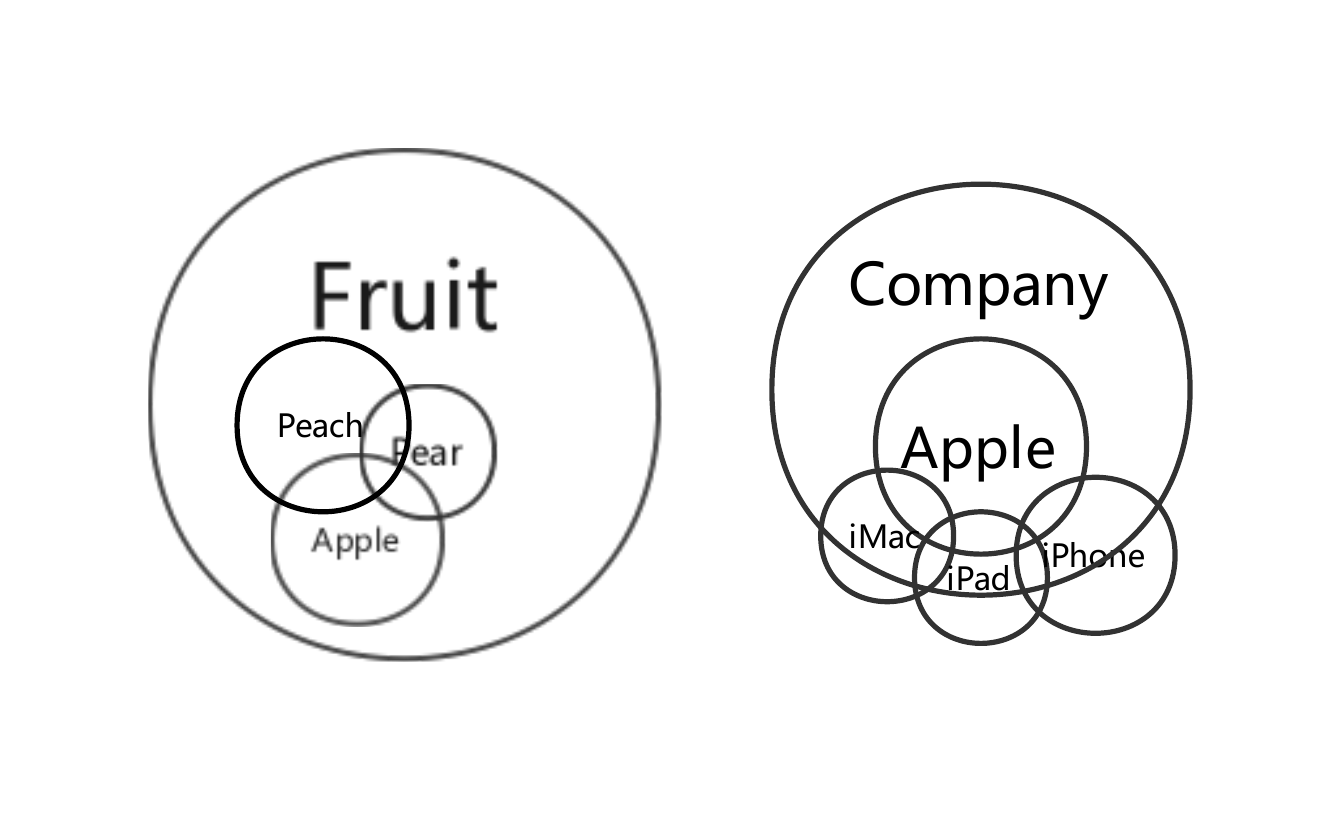}
  \caption{An example of multi-sense words}
\end{figure}

Maximum cosine similarity
\begin{equation}
    d(w_i, w_j) =\max_{p,q=1,...,K}\frac{\langle\mu_{i,p},\mu_{j,q} \rangle}{\|\mu_{i,p} \|\cdot \| \mu_{j,q}\|}
\end{equation}
where $p$,$q$ are the index of the Gaussian distribution in its Gaussian mixture. Note the cosine similarity is equivalent to the normalized Euclidean distance $\|\cdot\|_2$. We use this metric to evaluate the models.

Expected likelihood \cite{jebara2004probability}:
\begin{equation}
\begin{split}
    E(f(w_i), f(w_j)) &= \\
    \sum_{p=1}^K\sum_{q=1}^K p_{i,p}p_{j,q}\int &\mathcal{N}(x;\mu_{i,p};\Sigma_{i,p})\mathcal{N}(x;\mu_{j,q};\Sigma_{j,q}) \mathrm{d}x  \\
    = \sum_{p=1}^K \sum_{q=1}^K p_{i,p} q_{j,q}& \mathcal{N}(0;\mu_{i,p}-\mu_{j,q};\Sigma_{i,p}+\Sigma_{j,q})
\end{split}
\end{equation}

KL-divergence:
\begin{equation}
\begin{split}
    D_{KL}(f(w_{i,p})\| f(w_{j,q})) &= \\
    \int \mathcal{N}(x;\mu_{j,q};\Sigma_{j,q})&\log \frac{\mathcal{N}(x;\mu_{i,p};\Sigma_{i,p})}{\mathcal{N}(x;\mu_{j,q};\Sigma_{j,q})} \mathrm{d}x\\
    =\frac{1}{2}(d + \log\frac{\mathrm{det}(\Sigma_{i,p})}{\mathrm{det}(\Sigma_{j,q})}&-\mathrm{tr}(\Sigma_j^{-1}\Sigma_i
    \\
    -(\mu_{j,q}-\mu_{i,p})\Sigma_{i,p}^{-1}(&\mu_{j,q}-\mu_{i,p}) )
\end{split}
\end{equation}
Note that KL-divergence is an asymmetric measure of similarity between two distributions.

\subsection{Training}
We used similar hyper-parameters in Word2Vec and Word2GM training, in order to make the results comparable with each other. Both models applied skip-gram context window $\ell=5 $, space dimensions $D=50$ and only contained words with word frequency $\geq 5$. Training a Word2Vec model is fairly straightforward.

It is yet unclear how to properly choose the number of Gaussians per word $K$, the boundaries of $\|\mu \|_2$ and $\Sigma$ and their initial values, thus we simply borrowed the model from the original paper of Word2GM \cite{athiwilson2017}. During training, we observed that the vector space is dense in terms of the radius $\max(\|\mu_i\|)$ and covariance matrices $\Sigma$, and it might affect the convergence speed of the adaptive gradient descent algorithm we used. 

\subsection{Testing}
We examined the task performance on several models. Firstly, we trained a Word2Vec model with 250M English Wikipedia corpus. Secondly, we trained a Word2GM model with the same Wikipedia corpus and a Word2GM model with a trimmed 100M English Wikipedia corpus since training Word2GM models is quite time-consuming and memory-expensive, and we're currently investigating how to optimize the Gaussian mixture embedding algorithm. Finally, we compared these models with a pre-trained Word2Vec model from Facebook FastText.  
\vspace{12pt}

\section{Evaluation}
\vspace{12pt}

\begin{table}[htb]
\begin{center}
\begin{tabular}{|l|l|l|}
\hline \bf Model & \bf Accuracy&\bf Score\\ \hline
w2v-250m & 0.64 &\textbf{1.55} \\
w2v-fb & \bf 0.68 &1.66  \\
w2gm-250m & 0.54 & 2.09 \\
w2gm-100m & 0.50 &2.25 \\
\hline
\end{tabular}
\end{center}
\caption{\label{font-table} Task 1 Evaluation }
\end{table}

\begin{table}[htb]
\begin{center}
\begin{tabular}{|l|l|}
\hline \bf Model & \bf Score\\ \hline
w2v-250m & \textbf{4.8\%} \\
w2v-fb & 3.0\% \\
w2gm-250m & 3.2\%\\
w2gm-100m & 2.2\% \\
\hline
\end{tabular}
\end{center}
\caption{\label{font-table}Task 2 Evaluation }
\end{table}

As shown in Table 1 and Table 2 the Word2Vec models generally out-perform Word2GM models. The accuracy was calculated based on the best 1 prediction per OOV word while the score takes all the predictions into consideration. In Task 2, only the 5 best predictions for each OOV word were scored since making more prediction attempts would drop the scores. The result of Task 2 indicates that the unsupervised model outputs are still far from how humans use English. Therefore, we want these results to become the baseline of the tasks. 
        
\vspace{12pt}

\section{Conclusions}
\vspace{12pt}

This is the first attempt to address the issues of OOV lexical prediction in NLP tasks with unstructured data and given no prior knowledge. Hence, we propose two tasks for OOV classification and prediction, then we create baseline results with several models based on Word2Vec and Word2GM algorithms. Our result has shown that, OOV lexical prediction is still challenging with unsupervised word embedding models.

\vspace{12pt}

\bibliographystyle{IEEEbib}
\vspace{12pt}

\bibliography{strings}

\begin{thebibliography}{10}

\bibitem{zhang2015zero}
Ziming Zhang and Venkatesh Saligrama,
\newblock ``Zero-shot learning via semantic similarity embedding,''
\newblock in {\em Proceedings of the IEEE International Conference on Computer
  Vision}, 2015, pp. 4166--4174.

\bibitem{wang2016self}
Ye~Wang and Mi~Lu,
\newblock ``A self-adaptive algorithm to defeat text-based
  \uppercase{CAPTCHA},''
\newblock in {\em Industrial Technology (ICIT), 2016 IEEE International
  Conference on}. IEEE, 2016, pp. 720--725.

\bibitem{wang2017combining}
Ye~Wang, Yuanjiang Huang, Wu~Zheng, Zhi Zhou, Debin Liu, and Mi~Lu,
\newblock ``Combining convolutional neural network and self-adaptive algorithm
  to defeat synthetic multi-digit text-based \uppercase{CAPTCHA},''
\newblock in {\em Industrial Technology (ICIT), 2017 IEEE International
  Conference on}. IEEE, 2017, pp. 980--985.

\bibitem{siddharthan2006syntactic}
Advaith Siddharthan,
\newblock ``Syntactic simplification and text cohesion,''
\newblock {\em Research on Language \& Computation}, vol. 4, no. 1, pp.
  77--109, 2006.

\bibitem{zhang2016effective}
Yue Zhang and Xinxiang Zhang,
\newblock ``effective real-scenario video copy detection,''
\newblock in {\em 2016 23rd International Conference on Pattern Recognition
  (ICPR)}. IEEE, 2016, pp. 3951--3956.

\bibitem{Yin2019SpatialTemporal}
Zhengcong Yin, Chong Zhang, Daniel~W Goldberg, and Sathya Prasad,
\newblock ``An nlp-based question answeringframework for spatio-temporal
  analysis and visualization,''
\newblock in {\em ICGDA '19: Proceedings of the International Conference on
  Geoinformatics and Data Analysis}, New York, NY, USA, 2019, ACM.

\bibitem{dagan2006pascal}
Ido Dagan, Oren Glickman, and Bernardo Magnini,
\newblock ``The pascal recognising textual entailment challenge,''
\newblock in {\em Machine learning challenges. evaluating predictive
  uncertainty, visual object classification, and recognising tectual
  entailment}, pp. 177--190. Springer, 2006.

\bibitem{curran2005supersense}
James~R Curran,
\newblock ``Supersense tagging of unknown nouns using semantic similarity,''
\newblock in {\em Proceedings of the 43rd Annual Meeting on Association for
  Computational Linguistics}. Association for Computational Linguistics, 2005,
  pp. 26--33.

\bibitem{ciaramita2003supersense}
Massimiliano Ciaramita and Mark Johnson,
\newblock ``Supersense tagging of unknown nouns in wordnet,''
\newblock in {\em Proceedings of the 2003 conference on Empirical methods in
  natural language processing}. Association for Computational Linguistics,
  2003, pp. 168--175.

\bibitem{biran2011putting}
Or~Biran, Samuel Brody, and No{\'e}mie Elhadad,
\newblock ``Putting it simply: a context-aware approach to lexical
  simplification,''
\newblock in {\em Proceedings of the 49th Annual Meeting of the Association for
  Computational Linguistics: Human Language Technologies: short papers-Volume
  2}. Association for Computational Linguistics, 2011, pp. 496--501.

\bibitem{specia2012semeval}
Lucia Specia, Sujay~Kumar Jauhar, and Rada Mihalcea,
\newblock ``Semeval-2012 task 1: English lexical simplification,''
\newblock in {\em Proceedings of the First Joint Conference on Lexical and
  Computational Semantics-Volume 1: Proceedings of the main conference and the
  shared task, and Volume 2: Proceedings of the Sixth International Workshop on
  Semantic Evaluation}. Association for Computational Linguistics, 2012, pp.
  347--355.

\bibitem{mccarthy2007semeval}
Diana McCarthy and Roberto Navigli,
\newblock ``Semeval-2007 task 10: English lexical substitution task,''
\newblock in {\em Proceedings of the 4th International Workshop on Semantic
  Evaluations}. Association for Computational Linguistics, 2007, pp. 48--53.

\bibitem{manning2008boolean}
Christopher~D Manning, Prabhakar Raghavan, and Hinrich Sch{\"u}tze,
\newblock ``Boolean retrieval,''
\newblock {\em Introduction to information retrieval}, pp. 1--18, 2008.

\bibitem{rumelhart1986david}
DE~Rumelhart,
\newblock ``David e. rumelhart, geoffrey e. hinton, and ronald j. williams,''
\newblock {\em Nature}, vol. 323, pp. 533--536, 1986.

\bibitem{bengio2003neural}
Yoshua Bengio, R{\'e}jean Ducharme, Pascal Vincent, and Christian Jauvin,
\newblock ``A neural probabilistic language model,''
\newblock {\em Journal of machine learning research}, vol. 3, no. Feb, pp.
  1137--1155, 2003.

\bibitem{mikolov2013distributed}
Tomas Mikolov, Ilya Sutskever, Kai Chen, Greg~S Corrado, and Jeff Dean,
\newblock ``Distributed representations of words and phrases and their
  compositionality,''
\newblock in {\em Advances in neural information processing systems}, 2013, pp.
  3111--3119.

\bibitem{wang2017comparisons}
Ye~Wang, Zhi Zhou, Shan Jin, Debin Liu, and Mi~Lu,
\newblock ``Comparisons and selections of features and classifiers for short
  text classification,''
\newblock in {\em IOP Conference Series: Materials Science and Engineering}.
  IOP Publishing, 2017, vol. 261, p. 012018.

\bibitem{reisinger2010multi}
Joseph Reisinger and Raymond~J Mooney,
\newblock ``Multi-prototype vector-space models of word meaning,''
\newblock in {\em Human Language Technologies: The 2010 Annual Conference of
  the North American Chapter of the Association for Computational Linguistics}.
  Association for Computational Linguistics, 2010, pp. 109--117.

\bibitem{huang2012improving}
Eric~H Huang, Richard Socher, Christopher~D Manning, and Andrew~Y Ng,
\newblock ``Improving word representations via global context and multiple word
  prototypes,''
\newblock in {\em Proceedings of the 50th Annual Meeting of the Association for
  Computational Linguistics: Long Papers-Volume 1}. Association for
  Computational Linguistics, 2012, pp. 873--882.

\bibitem{tian2014probabilistic}
Fei Tian, Hanjun Dai, Jiang Bian, Bin Gao, Rui Zhang, Enhong Chen, and Tie-Yan
  Liu,
\newblock ``A probabilistic model for learning multi-prototype word
  embeddings.,''
\newblock in {\em COLING}, 2014, pp. 151--160.

\bibitem{vilnis2014word}
Luke Vilnis and Andrew McCallum,
\newblock ``Word representations via gaussian embedding,''
\newblock {\em arXiv preprint arXiv:1412.6623}, 2014.

\bibitem{wang2019attention}
Ye~Wang, Han Wang, Xinxiang Zhang, Theodora Chaspari, Yoonsuck Choe, and Mi~Lu,
\newblock ``An attention-aware bidirectional multi-residual recurrent neural
  network (abmrnn): A study about better short-term text classification,''
\newblock in {\em ICASSP 2019-2019 IEEE International Conference on Acoustics,
  Speech and Signal Processing (ICASSP)}. IEEE, 2019, pp. 3582--3586.

\bibitem{yin2019DLRG}
Zhengcong Yin, Andong Ma, and Daniel~W Goldberg,
\newblock ``A deep learning approach for rooftop geocoding,''
\newblock {\em Transactions in GIS}, 2019.

\bibitem{wang2018optimized}
Ye~Wang and Mi~Lu,
\newblock ``An \uppercase{O}ptimized \uppercase{S}ystem to \uppercase{S}olve
  \uppercase{T}ext-based \uppercase{CAPTCHA},''
\newblock {\em arXiv preprint arXiv:1806.07202}, 2018.

\bibitem{wu2019accurate}
Hao Wu, Xinxiang Zhang, Brett Story, and Dinesh Rajan,
\newblock ``Accurate vehicle detection using multi-camera data fusion and
  machine learning,''
\newblock in {\em ICASSP 2019-2019 IEEE International Conference on Acoustics,
  Speech and Signal Processing (ICASSP)}. IEEE, 2019, pp. 3767--3771.

\bibitem{Zhang2019Visual}
Chong Zhang, Zhengcong Yin, Peng Gao, and Sathya Prasad,
\newblock ``A visual analytics approach to explorationof hotels in overlaid
  drive-time polygons of attractions,''
\newblock in {\em International Symposium On Web and Wireless Geographical
  Information Systems}. 2019, Springer International Publishing.

\bibitem{athiwilson2017}
Ben Athiwaratkun and Andrew~Gordon Wilson,
\newblock ``Multimodal word distributions,''
\newblock in {\em Conference of the Association for Computational Linguistics
  (ACL)}, 2017.

\bibitem{goldberg2014word2vec}
Yoav Goldberg and Omer Levy,
\newblock ``Word2vec explained: Deriving mikolov et al.'s negative-sampling
  word-embedding method,''
\newblock 2014.

\bibitem{jebara2004probability}
Tony Jebara, Risi Kondor, and Andrew Howard,
\newblock ``Probability product kernels,''
\newblock {\em Journal of Machine Learning Research}, vol. 5, no. Jul, pp.
  819--844, 2004.

\end{thebibliography}

\end{document}